# PAEDID: Patch Autoencoder Based Deep Image Decomposition For Pixel-level Defective Region Segmentation

Shancong Mou[1], Meng Cao[2], Haoping Bai[2], Ping Huang[2], Jianjun Shi[1*] and Jiulong Shan[2]

[1]*H. Milton Stewart School of Industrial and Systems Engineering, Georgia Institute of Technology, Atlanta, GA, USA.*

[2]*Apple, Cupertino, CA, USA.*

## Abstract

Unsupervised pixel-level defective region segmentation is an important task in image-based anomaly detection for various industrial applications. The state-of-the-art methods have their own advantages and limitations: matrix-decomposition-based methods are robust to noise but lack complex background image modeling capability; representation-based methods are good at defective region localization but lack accuracy in defective region shape contour extraction; reconstruction-based methods detected defective region match well with the ground truth defective region shape contour but are noisy. To combine the best of both worlds, we present an unsupervised patch autoencoder based deep image decomposition (PAEDID) method for defective region segmentation. In the training stage, we learn the common background as a deep image prior by a patch autoencoder (PAE) network. In the inference stage, we formulate anomaly detection as an image decomposition problem with the deep image prior and sparsity regularizations. By adopting the proposed approach, the defective regions in the image can be accurately extracted in an unsupervised fashion. We demonstrate the effectiveness of the PAEDID method in simulation studies and an industrial dataset in the case study.

## 1. Introduction

### *1.1 motivation*

Image-based anomaly detection is an important task in various industrial and medical applications (Yan et al. 2017). It can be categorized into the following two categories: (i) from a supervision point of view: supervised and unsupervised anomaly detection. (Pang et al. 2021) (ii) from task point of view: instance-level (i.e., identify anomaly samples (Gong et al. 2019, Wang et al. 2021)), localization-level (i.e., localize the defect (Bergmann et al. 2020, Wang et al. 2021)), and pixel-level (i.e., extract pixel-wise defect contour (Minaee et al. 2015, Yan et al. 2017, Zhou and Paffenroth 2017, Mou et al. 2021)).

Supervised anomaly detection methods require sufficient labeled training samples such as instance-level labels for identifying anomaly samples, localization-level labels for object detection (Liu et al. 2020), or pixel-level labeled images for segmentation (Minaee et al. 2021), which are expensive and time-consuming to collect in manufacturing processes. Notice that, instance-level anomaly detection can be seen as a binary classification task, where the class activation map from the classifier can provide a rough defective region localization mask even without localization-level labels (Minaee et al. 2021).

Unsupervised image-based anomaly detection gains popularity in various manufacturing processes since it avoids the labeling process (Yan et al. 2017). Among all those tasks, unsupervised pixel-level defective region segmentation can provide fine-scale defect information such as defect shape contour, aspect ratio and so on. That information is important for reducing the false alarm rate for defect type classification, accurate annotation, and so on. It is also one of the most challenging tasks since the defect detail needs to be inferred from the image data in an unsupervised fashion.

### *1.2 Literature review*

Multiple methods have been proposed for unsupervised defective region segmentation, including matrix-decomposition-based and deep-learning-based methods. Among them, deep-learning-based methods can be further categorized as deep-representation-based and deep-reconstruction-based methods (Yu et al. 2021). A detailed review of these methods is introduced as follows.

### *1.2.1 Matrix-decomposition-based methods*

To achieve accurate unsupervised defective region segmentation, one feasible solution is to use matrix-decomposition-based methods (Xu et al. 2010, Candès et al. 2011, Peng et al. 2012, Mardani et al. 2013, Minaee et al. 2015, Yan et al. 2017, Yan et al. 2018, Mou et al. 2021, Mou and Shi 2022). In these methods, statistical priors (i.e., smoothness, low-rank of the background, and sparsity, basis representation of the anomalies) are incorporated as regularization terms in an optimization problem to achieve the background restoration and defective region segmentation. An advantage of the matrix-decomposition-based method is its robustness, with respect to anomalous region, when restoring the background from the test image corrupted by defective regions.

The commonly used matrix decomposition methods include (i) Robust principal component analysis (RPCA) type of algorithms (Xu et al. 2010, Candès et al. 2011, Peng et al. 2012, Mardani et al. 2013), which decompose a data matrix into a low-rank matrix of backgrounds and a sparse matrix of anomalies; and (ii) Smooth sparse decomposition (SSD) type of algorithms (Minaee et al. 2015, Yan et al. 2017, Yan et al. 2018, Mou et al. 2021, Mou and Shi 2022), which decompose an image into a smooth background and the sparse anomalies. RPCA was first proposed by Candès et al. (2011) as a modification of PCA (Friedman 2017) to enhance its robustness for grossly corrupted observations. Since then, it has been widely used in video surveillance applications, where the data matrix is assembled by vectorized images with a stationary (or slowly varying) background and a sparse foreground (Bouwmans and Zahzah 2014). Outlier Pursuit (Xu et al. 2012) aims to decompose the data matrix into low-rank and column-wise sparse components, which can be used for image-level anomaly detection. The low-rank assumption requires perfect alignment across multiple images which is seldom satisfied in real-world applications. To address the linear misalignment issue, Peng et al. (2012) generalized the RPCA method to RASL by conducting linear alignment and decomposition at the same time. SSD was first proposed for anomaly detection in images with smooth background and sparse anomalies (Yan et al. 2017) and then generalized to spatiotemporal data (Yan et al. 2018). Mou et al. (2021) generalized matrix decomposition methods to Additive Tensor Decomposition (ATD) framework, which can deal with images with smoothness, low-rank, or piecewise constancy properties.

Matrix-decomposition-based methods have successful applications such as indentation region extraction on a silicon surface (Yan et al. 2017), calcification region extraction in the medical images (Mou et al. 2021), defective region extraction in misaligned images (Peng et al. 2012). However, statistical priors are less capable of modeling complex backgrounds that do not have smooth or low-rank properties, which is common in manufacturing applications. For example, Figure 1 shows the images of cross-sections of the wood structure without defects. Inside the same image, it has a unique structure that cannot be described with any domain regularization type of priors. Across multiple images, there are similar structures, but the matrix composed of vectorized images is not low-rank, even after alignment. This special, but common, structure in a manufacturing process raises challenges for state-of-the-art matrix-decomposition-based anomaly detection algorithms.

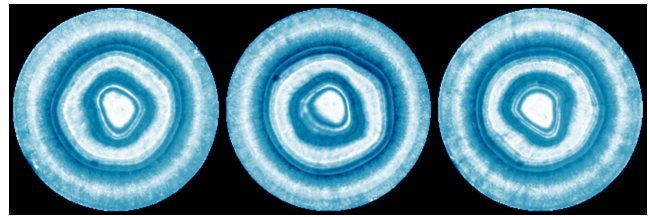

(a) Normal images of product 1

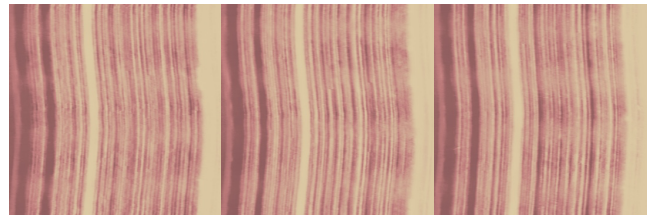

(b) Normal images of product 2

Fig. 1. Normal images with complex backgrounds from BTAD dataset (Mishra et al. 2021)

*1.2.2 Deep-reconstruction-based methods*

Recently, deep-learning-based methods receive more and more attention in unsupervised anomaly detection due to their superior modeling capability in complex images. Deep-reconstruction-based methods usually reconstruct the normal background of input anomalous image from a learned subspace and assume such a subspace does not generalize to anomalies to ensure its robustness. Then the reconstruction residual is used for extracting a defective region segmentation map. The commonly used methods in learning the low dimensional space and distribution of normal images are autoencoders (AEs) (Zhou and Paffenroth 2017, Bergmann et al. 2018, Gong et al. 2019) and generative adversarial networks (GANs) (Schlegl et al. 2017, Zenati et al. 2018, Schlegl et al. 2019), respectively. For example, Bergmann et al. (2018) trained a structural similarity loss-based AE (SSIM-AE) for defective region segmentation. Schlegl et al. (2017) trained a GAN on normal images and used the pixel-wise reconstruction error for anomalous regions localization. Such a learned subspace can serve as a prior to model various backgrounds. However, a common challenge of those methods lies in balancing the generalization capability and modeling capability: a subspace that achieves a satisfactory reconstruction of normal regions of the input image also "generalizes" so well that it can always reconstruct the abnormal inputs as well (Gong et al. 2019). This usually leads to either a noisy residual map (segmentation mask) due to the lack of background reconstruction capability or missing detection of anomalous pixels due to the overfitting of the reconstructed image towards the anomalous input. Several methods (Zhou and Paffenroth 2017, Gong et al. 2019) are proposed to restrict such generalization capability.

Recently, methods based on AE variations are also used for anomaly detection, including variational AEs (An and Cho 2015) and AE trained with adversarial loss (Akcay et al. 2018). VAE has poor performance even in a relatively unchallenging dataset (Akcay et al. 2018). Moreover, the image reconstructed from VAE tends to be blurry (Zhao et al. 2017), which hinders its application in pixel-wise anomaly detection since any mismatch between the restored image and the input image will be counted towards the anomaly score. GANomaly (Akcay et al. 2018) adopts AE architecture comprising an adversarial training module for anomaly detection. However, a pixel-wisely accurate background reconstruction which is necessary for a good performance in pixel-wise anomaly detection, is still challenging for most of those methods.

Another closely related field is the open-set semantic segmentation (Cen et al. 2021), which aims to identify out-of-distribution pixels among all pixels in one image. It is an important task in autonomous driving applications. When there is only one class of in-distribution objects, i.e., the normal background, open-set semantic segmentation is also called anomaly segmentation. Therefore, it shares similar approaches with pixel-wise anomaly detection methods reviewed thereof, including AE-based methods (Creusot and Munawar 2015, Baur et al. 2018) and GAN-based methods (Lis et al. 2019, Bevandić et al. 2022).

*1.2.3 Deep-representation-based methods*

Deep representation-based methods learn the discriminate embeddings of normal images from a clean training set and achieve anomaly detection by comparing the embedding of a test image and the distribution of the normal image embeddings (Defard et al. 2021, Roth et al. 2021, Wang et al. 2021a, Gudovskiy et al. 2022). These methods usually use pre-trained feature extractors (Roth et al. 2021) on large-scale natural image datasets (i.e., ImageNet (Deng et al. 2009)) and have the advantage when the training samples are limited. However, modern manufacturing processes usually are data-rich environments (even though label-rare) where collecting training samples are not difficult. While pretrained feature extractors are good at localizing the defective region, they lack the capacity to generate pixel-level accurate masks or contours. For example, Roth et al. (2021) proposed the PatchCore method to directly use the output of the first several layers of the ImageNet pre-trained network on normal images to extract features to form a memory bank. For any test image, the anomaly score for defective region localization can be calculated for each patch by comparing the features with the features in the memory bank. Despite the locally aware property of the PatchCore approach, it loses resolution for accurate defective region segmentation.

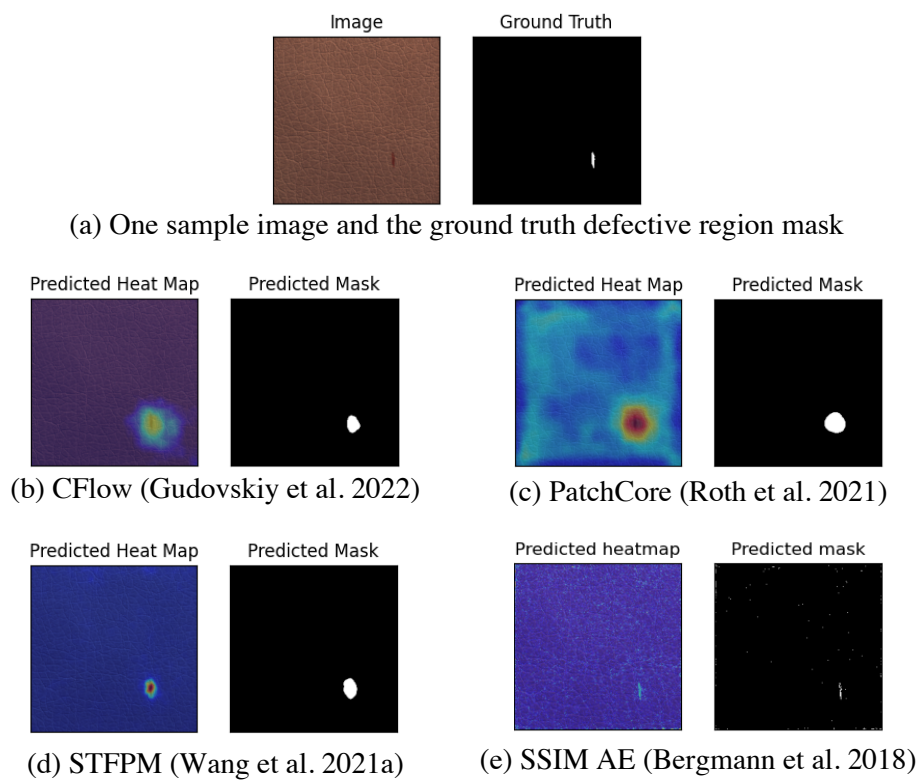

(a) One sample image and the ground truth defective region mask

(b) CFlow (Gudovskiy et al. 2022)

(c) PatchCore (Roth et al. 2021)

(d) STFPM (Wang et al. 2021a)

(e) SSIM AE (Bergmann et al. 2018)

Fig. 2. Sample defective region segmentation results: (a) one sample image and ground truth defective region mask; (b-d) results of three state-of-the-art methods (Akcay et al. 2022); (e) result of an SSIM AE (Bergmann et al. 2018)

Figure 2 (a-c) shows the results of multiple state-of-the-art representation-based methods in an example anomaly image of the MVTec (Bergmann et al. 2019) dataset. Even though all the methods can highlight pixels that are in the general vicinity of the defective region, their outputs cannot provide a precise contour around the defect. Therefore, there is a research gap in generating pixel-accurate defective region segmentation masks.

In summary, matrix-decomposition-based methods with statistical priors admit robustness but lack complex background image modeling capability. The learned subspace from reconstruction-based methods can serve as a prior to model various backgrounds but can potentially overfit the defective regions. Representation-based methods are good at defective region localization but lack accuracy in extracting the shape contour of the defective region. Therefore, a method that combines the best of both worlds would be ideal.

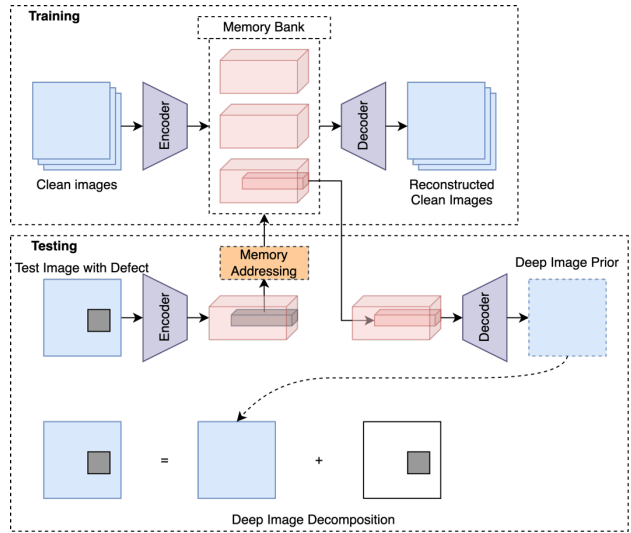

Fig. 3. Overview of the proposed PAEDID framework

*1.3 Proposed method and contributions*

In this paper, we propose a Patch Autoencoder based Deep Image Decomposition (PAEDID) framework (Figure 3) that adapts to the data-rich but label-rare manufacturing environment. To fully utilize the information in the rich normal data to enhance the background modeling capability, the normal background is modelled as a deep prior learned from normal images. Moreover, a local patch-based AE is proposed to restrict the generalizability of the deep neural net and localize the defective region better, while maintaining its capability of modeling complex image distributions. Then, a penalized optimization problem is solved to decompose the image into the background and defective region. Intuitively, the proposed PAEDID method inherits the robustness from matrix decomposition methods and the learned prior through deep-reconstruction-based methods enables handling complex backgrounds beyond traditional statistical priors.

Our contributions can be summarized in three aspects: (i) We propose a novel deep image decomposition framework for unsupervised pixel-level defective region segmentation framework that outperforms existing methods in terms of defective region segmentation. The framework does not require pixel-wise labeled training samples, which is suitable for defective region segmentation tasks in data-rich but label-rare manufacturing processes. (ii) The proposed method integrates the modeling capability of reconstruction-based methods in complex backgrounds and the robustness and data efficiency of matrix decomposition in defective region modeling. (iii) Our framework is general and extendable. The deep image prior can be replaced with various generative models (e.g., GANs and AE variants) depending on the tasks and we leave that for future work.

The remainder of this paper is organized as follows. The PAEDID framework is presented in Section 2. Section 3 presents several simulation studies based on the PAEDID framework. In Section 4, a case study on solar cell defective region segmentation application is utilized to demonstrate the effectiveness of the proposed framework. Section 5 concludes the paper.

## 2. Patch Autoencoder based Deep Image Decomposition

In this section, we will introduce the proposed PAEDID method. As shown in Figure 3, we propose a two-step method. First, a patch-based AE is trained on a set of clean images to construct a memory bank of latent representations of normal images. Then, the abnormal patches with the top $\alpha$ percentage anomaly score in the latent representation of a test image are replaced by the corresponding normal patches in the memory bank. Next, the updated latent representation is used to reconstruct the deep image prior that describes the corresponding normal background. Finally, an optimization problem with the deep image prior and sparsity regularizations is formulated to decompose the test image into the normal background and defective region. The main assumption of the proposed method is that we have a relatively large training set that covers all possible variations of clean images. In the rest of this section, we first discuss how to learn the distribution of normal images, and then introduce the way to extract anomalies from a test image in the inference stage.

## 2.1 Learning normal image distribution using AEs

This section corresponds to the training stage in Figure 3. The training stage includes the training of an AE network and the building of a memory bank consisting of latent representations of training images.

### 2.1.1 A brief introduction of AEs

AEs are widely used as an unsupervised nonlinear dimensional reduction tool in the deep learning context (Hinton and Salakhutdinov 2006). An AE consists of two components, an encoder network $E_{w_1}(\cdot)$, and a decoder network $D_{w_2}(\cdot)$, where $w_1$ and $w_2$ are the parameters of the encoder ($w_1$) and decoder ($w_2$). The encoder learns a mapping from the high dimensional input space to a low dimensional latent space; the decoder learns to reconstruct the input from its latent representation. Given $d_1 \times d_2$ normal images with $d_3$ channels, i.e., $\mathcal{X} = \{X_1, \ldots, X_n\}$ where $X_i \in R^{d_1 \times d_2 \times d_3}$, the AE learns the parameters $w_1$ and $w_2$ simultaneously by solving the following optimization problem:

$$\min_{w_1, w_2} \sum_{i=1}^{n} \left\| vec\left(X_i - D_{w_2}\left(E_{w_1}(X_i)\right)\right) \right\|_2^2 \tag{1}$$

where $vec(\cdot)$ is vectorization.

### 2.1.2 Patch-based AEs

As mentioned in Section 1, sometimes the trained AEs generalize so well such that they can reconstruct the defective regions well which significantly restricts their application in accurate defective region segmentation applications. To model such a complex background in a manufacturing process (e.g. Figure 2), the latent space of AE has to be large. In this case, inevitable generalization to the defective region could occur. For example, Figure 4 compares two AEs with different capability and their generalization behaviors on the same set of images. Figure 4 (a) shows an AE with a small latent space which does not generalize to defective regions but cannot model image detail. On the contrary, an AE with a large latent space (Figure 4 (b)) will perform better in the detailed modeling while generalizing well to defective regions. Therefore, restricting such generalization capability of AE while maintaining sufficient modeling capacity is essential. To achieve this goal, we adopt a patch-based AE which consists of three components: encoder, decoder, and a patch-based memory bank. Notice that the patch-based method has been utilized in various representation-based methods, such as PatchCore (Roth et al. 2021). However, those methods mainly focus on anomaly detection in a patch-based latent space, where the memory bank is used as a standard normal set for comparison. Even though similar concepts are utilized, we aim for a totally different goal that leads to different design characteristics. The patch-based AE is used to produce a deep image prior for the background, with great detail. Therefore, we prefer an AE model with a relatively large latent space that tends to overfit a normal image, and the memory bank here is utilized to restrict its generalization capability to unseen defects. The detail of the patch-based AE will be introduced in the following discussion.

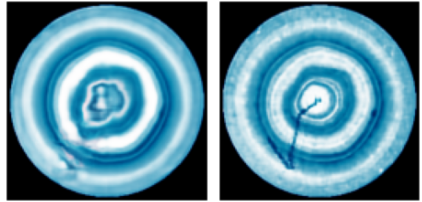

(a) $8 \times 8 \times 128$    (b) $32 \times 32 \times 128$

Fig. 4. Modeling and generalization performance of AEs with different latent space sizes

**Encoder and decoder.** The convolutional encoder and decoder are used to capture local information inside each image. Different from a vanilla AE (or regular patch-based AE), a relatively large latent space is adopted purposely to ensure its detail modeling capability.

Let $M_i = E(X_i)$ to be the feature map of input image $X_i$. Notice that $M_i \in \mathbb{R}^{P_1 \times P_2 \times P_3}$ is an order-3 tensor of height $P_1$, width $P_2$, and depth $P_3$. $M_i(p_1, p_2, :)$ is a $P_3$ dimensional vector representing the local feature at spatial location $p_1 \in \{1, ..., P_1\}$ and $p_2 \in \{1, ..., P_2\}$, which is a feature map corresponding to a specific patch of the original image. The receptive field size of the convolutional filter determines the local feature's representation area. It should be determined according to the local variation size of specific applications. Ideally, a balance between the receptive field size and the network depth should be considered such that the receptive field is large enough to account for local variations in the image while maintaining the modeling power.

The decoder has a similar structure as the encoder network but with reverse-ordered layers to reconstruct the input image. The encoder and decoder network are trained on a set of clean training images with the same loss function (1).

### 2.1.2 *Memory bank of patch features.*
After the encoder and decoder networks are trained, two memory banks are constructed:

(i) The raw memory bank, $\mathcal{M}_{raw} \in \mathbb{R}^{I \times P_1 \times P_2 \times P_3}$, which consist of the latent representations of each patch from all $I$ training images, i.e., $\mathcal{M}_{raw} = \{M_1, \ldots, M_I\}$ where $M_i$ is the latent representation of the $i$-th training image.

(ii) The aggregated memory bank of aggregation length $l$, $\mathcal{M}_{agg,l} \in \mathbb{R}^{I \times P_1 \times P_2 \times P_3\left(2\left\lfloor\frac{l}{2}\right\rfloor+1\right)^2}$, which consists of the aggregated latent representations of each patch from all training images (Roth et al. 2021), $\mathcal{M}_{agg,l} = \{M^1_{agg,l}, \ldots, M^I_{agg,l}\}$. The aggregation process is as follows: For the $(p_1, p_2)$-th patch inside the $i$-th training image, we collect the neighboring $\lfloor l/2 \rfloor$ latent representations centered at the $(p_1, p_2)$--th location, i.e.,

$$M^i_{agg,l}(p_1, p_2, :) = \text{vec}\left(M_i\left(p_1 - \left\lfloor\frac{l}{2}\right\rfloor : p_1 + \left\lfloor\frac{l}{2}\right\rfloor, p_2 - \left\lfloor\frac{l}{2}\right\rfloor : p_2 + \left\lfloor\frac{l}{2}\right\rfloor, :\right)\right)$$

$$\in \mathbb{R}^{P_3\left(2\left\lfloor\frac{l}{2}\right\rfloor+1\right) \times \left(2\left\lfloor\frac{l}{2}\right\rfloor+1\right)}.$$

The aggregated memory bank is used for robust memory addressing which will be discussed in the following section.

(iii) We further metricize the raw memory bank (and the aggregated memory bank), $\mathcal{M}^{mat}_{raw} \in \mathbb{R}^{IP_1P_2 \times P_3}$ (and $\mathcal{M}^{mat}_{agg,l} \in \mathbb{R}^{IP_1P_2 \times P_3\left(2\left\lfloor\frac{l}{2}\right\rfloor+1\right)^2}$), such that each row corresponding to one patch latent representation (one aggregated patch latent representation).

To further reduce the memory bank size, a coreset subsampling (Roth et al. 2021) step can be adopted. The algorithm can be found in Appendix C. Notice that the memory bank can be updated in an online fashion along with the manufacturing process.

## 2.2 Deep image decomposition

This section corresponds to the testing stage in Figure 3. The testing stage includes memory addressing and image decomposition.

### 2.2.1 Memory addressing

For any test image $X$, a query feature map $\widehat{M} = E(X)$ and the aggregated feature map $\widehat{M}_{agg,l}$ are generated. Then, for the $(p_1, p_2)$-th patch latent representation, $\widehat{M}(p_1, p_2, :)$, a search for the closest patch latent representation in the memory bank is performed. Motivated by Roth et al. (2021), we compared the neighborhood aggregated latent representation of this patch, $\widehat{M}_{agg,l}(p_1, p_2, :)$, with the aggregated latent representations in the aggregated memory bank $\mathcal{M}_{agg,l}^{mat}$, using the following distance measure: $d(M_1, M_2) = \frac{\langle M_1, M_2 \rangle}{\|M_1\|_2 \|M_2\|_2}$, where $M_1$ and $M_2$ are two latent representations to be compared. Denote the distance set to be $D\left(M_{agg,l}(p_1, p_2, :), \mathcal{M}_{agg,l}^{mat}\right) = \left\{d\left(M_{agg,l}(p_1, p_2, :), \mathcal{M}_{agg,l}^{mat}(j, :)\right) \mid j \in \{1, \ldots, IP_1P_2\}\right\}$. Denote the $k$-th smallest element in $D\left(M_{agg,l}(p_1, p_2, :), \mathcal{M}_{agg,l}^{mat}\right)$ to be $d_k$. Denote the index set of the $k$-nearest aggregated latent representations to be $C_{\mathcal{M}_{agg,l}^{mat}, k}\left(M_{agg,l}(p_1, p_2, :)\right) = \left(j \mid d\left(M_{agg,l}(p_1, p_2, :), \mathcal{M}_{agg,l}^{mat}(j, :)\right) \leq d_k\right)$. Then, the patches from the raw memory bank $\mathcal{M}_{raw}^{mat}$, corresponding to the $k$-nearest aggregated latent representations in $\mathcal{M}_{agg,l}$, are averaged to form a retrieved patch latent representation $\widehat{M}'(p_1, p_2, :)$, i.e.,

$$\widehat{M}'(p_1, p_2, :) = \frac{1}{k} \sum_{j \in C_{\mathcal{M}_{agg,l}^{mat}, k}\left(M_{agg,l}(p_1, p_2, :)\right)} \mathcal{M}_{raw}^{mat}(j, :).$$

Meanwhile, the corresponding anomaly score $s(p_1, p_2)$ can be calculated by averaging the distances between the neighborhood aggregated latent representation of the query patch and the $k$-nearest aggregated latent representation (Roth et al. 2021), i.e.,

$$s(p_1,p_2) = \frac{1}{k} \sum_{\substack{d \in D\left(M_{agg,l}(p_1,p_2,:), \mathcal{M}_{agg,l}^{mat}\right), \\ d \leq d_k}} d$$

The anomaly score $s(p_1, p_2)$ will have a large value if $\widehat{M}(p_1, p_2, :)$ cannot be approximated by any one of the feature maps in the memory bank of the same spatial location. In other words, $s$ indicates the defective region and thus can be used for defective region localization.

However, the anomaly score $s$ has the same width and height as the latent representation, which is not suitable for accurate defective region segmentation because of the low resolution. Therefore, a reconstruction step is needed. According to the assumption, the defective regions are sparse local regions. Therefore, in the reconstruction step, we only replace patches of the feature map with the top $\alpha$ anomalies scores, i.e.,

$$\widehat{M}''(p_1,p_2,:) = \begin{cases} \widehat{M}'(p_1,p_2,:), & \forall (p_1,p_2) \in A, \\ \widehat{M}(p_1,p_2,:), & otherwise, \end{cases}$$

where

$$A = \{(p_1,p_2) | s(p_1,p_2) > s_\alpha\},$$

and $s_\alpha$ is the $\alpha$ percentile value of $s$. $\alpha$ is determined by the engineering knowledge of the specific manufacturing process according to the size of the defective region.

The reconstructed image $D(\widehat{M}'')$ will serve as the deep image prior for the background. The algorithm of the deep image prior retrieval is listed in Algorithm 1.

**Algorithm 1** Algorithm for deep image prior retrieval

---
Training set $\mathcal{X} = \{X_1, \ldots, X_I\}$ of clean images; test image $X$ and parameter $\alpha$.
(1) AE network training:
  Train the encoder and the decoder simultaneously by solving the following optimization problem

$$\min_{w_1,w_2} \sum_{i=1}^{I} \left\| vec\left(X_i - D_{w_1}\left(E_{w_2}(X_i)\right)\right) \right\|_2^2.$$

(2) Memory banks construction

    a. Construct $\mathcal{M}_{raw}$

$$\mathcal{M}_{raw} = \{M_1, \ldots, M_I\} \in \mathbb{R}^{N \times P_1 \times P_2 \times P_3},$$

where

$$M_i = E_{w_2}(X_i), \quad \forall i \in \{1, \ldots, I\}.$$

    b. Construct $\mathcal{M}_{agg,l}$

$$\mathcal{M}_{agg,l} = \{M_{agg,l}^1, \ldots, M_{agg,l}^I\} \in \mathbb{R}^{n \times p_1 \times p_2 \times p_3\left(2\left\lfloor\frac{l}{2}\right\rfloor+1\right)^2},$$

where

$$M_{agg,l}^i = M_i\left(p_1 - \left\lfloor\frac{l}{2}\right\rfloor : p_1 + \left\lfloor\frac{l}{2}\right\rfloor, p_2 - \left\lfloor\frac{l}{2}\right\rfloor : p_2 + \left\lfloor\frac{l}{2}\right\rfloor, :\right).$$

    c. Matricize the raw memory bank (and the aggregated memory bank), $\mathcal{M}_{raw}^{mat} \in \mathbb{R}^{IP_1P_2 \times P_3}$ (and $\mathcal{M}_{agg,l}^{mat} \in \mathbb{R}^{IP_1P_2 \times P_3\left(2\left\lfloor\frac{l}{2}\right\rfloor+1\right)^2}$).

(3) Memory addressing:

For any test image $X$, generate a query feature map by $\widehat{M} = E(X)$ as well as the aggregated feature map $\widehat{M}_{agg,l}$.

**for** $p_1 \in \{1, \ldots, P_1\}$ **do**

  **for** $p_2 \in \{1, \ldots, P_2\}$ **do**

    a. Calculate the distance set:

$$D\left(M_{agg,l}(p_1, p_2, :), \mathcal{M}_{agg,l}^{mat}\right)$$
$$= \left\{d\left(M_{agg,l}(p_1, p_2, :), \mathcal{M}_{agg,l}^{mat}(j, :)\right) \mid j \in \{1, \ldots, IP_1P_2\}\right\}$$

    b. Find the index set of the $k$-nearest aggregated latent representations: $C_{\mathcal{M}_{agg,l}^{mat}, k}\left(M_{agg,l}(p_1, p_2, :)\right)$

    c. Calculate the retrieved patch latent representation for each patch of the encoded test image $\widehat{M} = E_{w_2}(X)$:

$$\widehat{M}'(p_1, p_2, :) = \frac{1}{k} \sum_{j \in C_{\mathcal{M}_{agg,l}^{mat}, k}\left(M_{agg,l}(p_1, p_2, :)\right)} \mathcal{M}_{raw}^{mat}(j, :),$$

    d. Calculate the anomaly score for each patch of the encoded test image $\widehat{M} = E_{w_2}(X)$.

$$s(p_1, p_2) = \frac{1}{k} \sum_{\substack{d \in D\left(M_{agg,l}(p_1, p_2, :), \mathcal{M}_{agg,l}^{mat}\right), \\ d \leq d_k}} d.$$

  **end**

**end**

(4) Feature map update and image reconstruction

    **for** $p_1 \in \{1, \ldots, P_1\}$ **do**

      **for** $p_2 \in \{1, \ldots, P_2\}$ **do**

$$\widehat{M}''(p_1, p_2, :) = \begin{cases} \widehat{M}'(p_1, p_2, :), & \forall (p_1, p_2) \in A, \\ \widehat{M}(p_1, p_2, :), & \text{otherwise}, \end{cases}$$

where

$$A = \{(p_1, p_2) \mid s(p_1, p_2) > s_\alpha\},$$

and $s_\alpha$ is the $\alpha$ percentile value of $s$.

|       **end**
|   **end**
The reconstructed image can be calculated as $\hat{X} = D(\widehat{M}'')$.

Notice that when images are aligned, which is also common in manufacturing inspection applications, the memory addressing algorithm can be further simplified by only searching for the closest patch in the memory bank of the same spatial location.

### 2.2.2 *Deep image decomposition*

In this section, we introduce the decomposition framework. We assume that any test image is a superimposition of (i) normal image (background $L$) from the distribution learned by the AE, and (ii) sparse defective regions ($S$), i.e., $X = L + S$. Then, the defective region segmentation can be formulated as a matrix decomposition problem with a combination of a deep image prior for the background and sparsity regularization for the defective regions. Given a test image $X$, we aim to extract the defective regions by solving the following penalized optimization problem:

$$\min_{L,S} l_{ssim}(L) + \lambda_1 l_{spar}(S) \qquad (2)$$

s.t.

$$X = L + S$$

where $\lambda_1$ is the tuning parameter; $l_{ssim}(L)$ is the similarity loss which evaluates the closeness of decomposed background $L$ and the reconstructed background $\hat{X}$ by the patch-based AE:

$$l_{ssim}(L) = \left\| L - \hat{X} \right\|_{SSIM},$$

where SSIM indicates the structural similarity loss (Wang et al. 2004) and a detailed introdcution of the SSIM loss can be found in Appendix A; $l_{spr}(S)$ denotes the sparsity penalty, i.e.,

$$l_{spar}(S) = \|S\|_1.$$

The optimization problem (2) can be converted to an unconstrained optimization problem

$$\min_L \left\| L - \hat{X} \right\|_{SSIM} + \lambda_1 \|X - L\|_1, \qquad (3)$$

and solved using an existing optimization solver, such as Adam (Kingma and Ba 2014) and its variants.

The proposed PAEDID method can be easily generalized to a noisy setting. We incorporate the problem formulation that explicitly considers measurement noise in Appendix B.

*2.3 Performance evaluation metric*

Multiple metrics in evaluating the defective region segmentation performance have been used in the existing literature, including the pixel-wise area under the receiver operator curve (pixel-wise AUROC score), per-region-overlap (PRO) (Bergmann et al. 2020), and the dice coefficient (Zou et al. 2004). The authors noticed that the pixel-wise AUROC (or PRO) may give misleading results in defective region segmentation tasks since those metrics are sensitive to class imbalance, i.e., the defective region only covers a small portion of pixels in the whole image. In this case, even a high pixel-wise AUROC score cannot guarantee a good pixel-level defective region segmentation performance. Various image-based anomaly detection algorithms (Bergmann et al. 2020, Defard et al. 2021, Roth et al. 2021, Wang et al. 2021a, Wang et al. 2021b, Yu et al. 2021) have demonstrated satisfactory performance in terms of pixel-wise AUROC score or PRO. Figure 2 (a-c) shows the results of multiple state-of-the-art methods on an example image of the leather defect category of the MVTec (Bergmann et al. 2019) dataset. Those methods showed unsatisfactory defective region segmentation results even with high pixel-wise AUROC scores. For example, the state-of-the-art PatchCore (Roth et al. 2021) method achieves an average pixel-wise AUROC score of 99.3% on the leather defect category of the MVTec (Bergmann et al. 2019) dataset, but the average best dice coefficient is 46.9%. Compared to the pixel-wise AUROC score and PRO, the dice coefficient is commonly used in semantic segmentation tasks which is more reliable in evaluating the defective region segmentation performance. Therefore, we use the dice coefficient which defined as

$$Dice(A, B) = \frac{2|A \cap B|}{|A + B|},$$

where $A$ is the predicted segmentation mask for the defective region and $B$ is the true segmentation mask. The dice coefficient evaluates the spatial overlap of the true segmentation mask and the predicted mask, and a larger dice coefficient indicates better performance of the detection algorithm.

### 2.4 Tuning parameter selection

In this section, we discuss the tuning parameter selection of the proposed method.

In general, there are two sets of hyper-parameters: (i) hyper-parameters related to AE network structure; (ii) hyper-parameters for the memory bank, including the aggregation length $l$ and nearest neighbor number $k$, and hyper-parameters for inference $\alpha$ and $\lambda_1$. Here, we refer to $l, k, \alpha$ and $\lambda_1$ as the tuning parameters of the proposed method.

The training of the AE network is independent of the deep image decomposition step, where the hyper-parameters are tuned on a training set of clean images, which follows the standard hyperparameter tuning procedure of AE networks (Bergmann et al. 2018).

Once the AE is trained, the memory bank can be established. Then, a deep image prior $\hat{X}$ for any input image $X$ can be retrieved following steps (3) and (4) in Algorithm 1. The retrieved deep image prior $\hat{X}$ is then incorporated in Eq (3) for image decomposition. In this procedure, tuning parameters $l, k, \alpha$, and $\lambda_1$ need to be specified. We present the following parameter tuning procedure similar to Mou et al. (2021): In phase 1, if there is training data $\{X_1, \ldots, X_m\}$ with pixel-wise annotated anomaly $\{S_1, \ldots, S_m\}$ available, the optimal tuning parameters can be determined by solving:

$$\hat{l}, \hat{k}, \widehat{\lambda_1}, \hat{\alpha} = \text{argmin}_{l,k,\lambda_1,\alpha} \sum_{i=1}^{m} \left\| X_i - \widehat{L_\iota}(l, k, \lambda, \alpha) - S_i \right\|_2,$$

where $\widehat{L_\iota}(l, k, \lambda, \alpha)$ is the estimated background by solving Eq (3). If there is no such annotation available, $l, k, \lambda$ and $\alpha$ should be selected empirically according to visual inspection of the decomposition quality.

### 2.5 Discussion

***The proposed PAEDID method is novel and outperforms the state of art methods, which can be justified with the following discussions:***

(i) Compared to traditional matrix-decomposition-based methods, the proposed PAEDID method can handle images with arbitrary backgrounds, in addition to those images following strict statistical priors. The background L in the proposed method is not assumed to be low-rank or smooth, but learned from training samples instead, which gives more flexibility in background modeling.

(ii) Compared to reconstruction-based methods, the proposed PAEDID method will give a cleaner detected defective region. Instead of directly using the reconstruction residual as a defective region indicator which will lead to a noisy segmentation mask, we use the reconstructed background as a deep image prior for image decomposition, the decomposed anomaly which admits sparsity property is much cleaner. We utilize the memory bank constructed from patches of normal images to restrict the AE's generalization capability while maintaining its modeling capability.

(iii) Compared to representation-based methods, segmentation masks obtained from the proposed PAEDID method have a higher resolution. Instead of localizing the defective region in the latent embedding space which will lose segmentation resolution, the anomaly score is utilized to identify anomalous patches. After replacing patches with a high normal score by corresponding normal ones from the memory bank, the re-assembled latent representation is decoded to serve as a deep image prior for the proposed pixel-wisely image decomposition algorithm. Therefore, the detected defective region has a higher resolution.

The superior performance of the proposed PAEDID method will be shown in the next section.

***The proposed PAEDID method is general and extendable.***

The major contribution of the proposed method is to use a learned prior to replace the highly restrictive statistical priors for background modeling in traditional matrix-decomposition methods, which makes it possible for pixel-wise sparse defective region detection in images with arbitrary backgrounds. Notice that in the current work, the deep image prior comes from a pretrained AE as well as a memory bank of latent representations of normal images. Such a deep image prior can be replaced with various generative models. For example, the generator from a pretrained GAN can also serve as such prior. We leave this for future work.

# 3. Simulation studies for performance evaluation

In this section, we use inspection images of two products of BTAD (Mishra et al. 2021) dataset to illustrate the PAEDID framework and demonstrate its effectiveness in defective region segmentation. Notice that the training set size in MVTec (Bergmann et al. 2019) is too small which doe not satisfy the assumption of the proposed method. For comparison, we also apply four representative state-of-the-art methods including SSIM AE (Bergmann et al. 2018), MemAE (Gong et al. 2019), PatchCore (Roth et al. 2021), and Robust Alignment by Sparse and Low-rank decomposition (RASL) (Peng et al. 2012). Among them, SSIM AE is the state-of-the-art vanilla AE-based benchmark method with the SSIM loss as the reconstruction loss; MemAE restricts the generalization capability of AE by an additional memory bank which demonstrated performance improvement in anomaly detection applications. PatchCore is the state-of-the-art representation-based anomaly detection method, which achieves superior performance in anomalous regions localization; RASL is the state-of-the-art matrix decomposition-based method, which can detect anomalies from slightly linear misaligned backgrounds. We assume that all input images are pre-identified defective images and evaluate the pixel-wise anomaly detection performance of the proposed method. The average dice coefficient (and standard deviation) of the proposed PAEDID method over all the test images on the test set is reported in Table 1.

Table 1. Average dice coefficient (and standard deviation) for different methods in simulation studies

|          | Product 1     | Product 2     |
|----------|---------------|---------------|
| SSIM AE  | 83.9% (5.1%)  | 81.8% (5.5%)  |
| MemAE    | 66.1% (7.7%)  | 73.8% (11.8%) |
| PatchCore| 22.5% (4.1%)  | 21.1% (3.5%)  |
| RASL     | 59.1% (10.1%) | 69.4% (11.2%) |
| PAEDID   | 98.9% (1.5%)  | 95.2% (2.8%)  |

## *3.1 Simulation study for Product 1*

For product 1, there are 1000 normal images of dimension 800×600 and we resize them into 128×128. 900 images are randomly selected as the training dataset in which 800 images are used as training and 100 images as the validation set to train the AE. The remaining 100 images are used to generate images with defects. The defects are simulated as cracks on top of the normal background, which are common in wood products. For example, Figure 5 (a) shows a simulated anomaly image with a corresponding normal background and generated defective region.

The encoder has a similar structure as the first several layers of VGG net (Simonyan and Zisserman 2014): Conv2D(128, 128, 64) - MaxPooling (2, 2) - ConvD2(64, 64, 128) - MaxPooling (2, 2). The decoder has the same structure but in reverse order. The latent space shape is $32 \times 32 \times 128$. After training, the memory bank is constructed by encoding the training dataset, where $l = 7, k = 13$. During testing, $\alpha = 0.3$.

Next, the reconstructed image is used as a deep image prior for decomposition. In this simulation study, the tuning parameter $\lambda = 0.00001$ is chosen by cross-validation which minimizes the dice coefficient. The detected defective regions are scaled to the range [0,1] and then a postprocessing step of global thresholding is applied to generate a binary defective region mask. The same postprocessing steps are applied for all methods to be compared. The threshold value is also chosen by

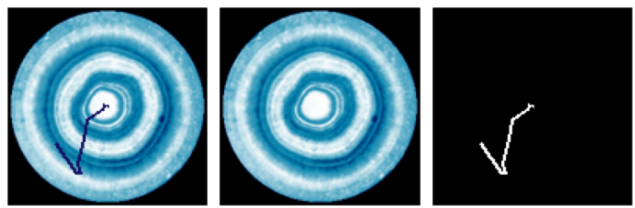

(a) Image with defect, true background, and simulated defective region

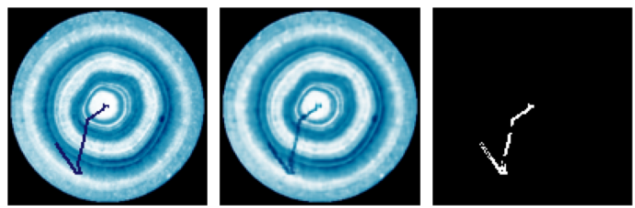

(b) Image with defect, SSIM AE reconstructed background and detected defective region

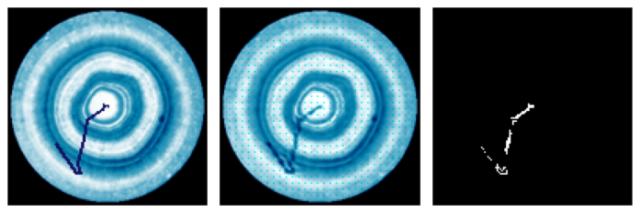

(c) Image with defect, MemAE reconstructed background and detected defective region

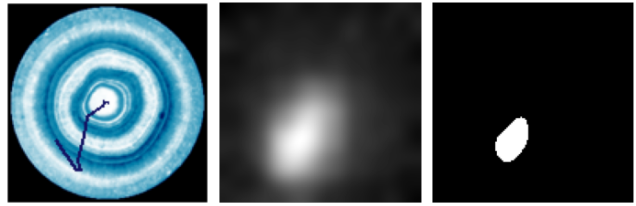

(d) Image with defect, PatchCore anomaly score map and detected defective region

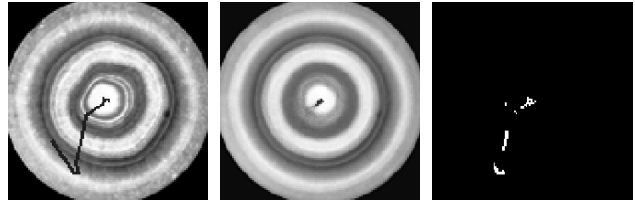

(e) Image with defect in greyscale, RASL reconstructed background and detected defective region

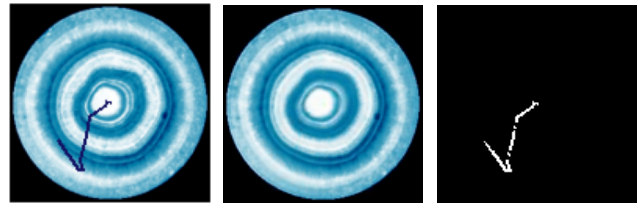

(f) Image with defect, PAEDID reconstructed background and detected defective region

Fig. 5. Comparison between the performance of SSIM AE, RASL PatchCore, and PAEDID cross-validation. Figure 5 shows the qualitative result of extracted defective regions of different methods.

The reconstructed background and detected defective region by SSIM AE (Bergmann et al. 2018) are shown in Figure 5 (b), which indicates that the extracted defective region is not satisfactory. This is because the generalization capability of the AE network is not restricted and will lead to a bad performance of defective extraction (Figure 5 (b)). The performance is also evaluated quantitatively in Table 1, where the dice coefficient is 83.9%.

The reconstructed background by MemAE (Gong et al. 2019) is shown in Figure 5 (c) where the extracted defective region is not clear. This is because MemAE restricted the generalization capability of the AE network by assigning a dictionary in the latent space, which deteriorates the modeling capability of AE. Therefore, the normal background cannot be reconstructed with high accuracy and will lead to a bad performance of defective region extraction (Figure 5 (c)). The performance is also evaluated quantitatively in Table 1, where the dice coefficient is 73.6%. We also notice that the training process of the MemAE is highly unstable, which may be due to training the additional dictionary in the latent space at the same time.

The PatchCore (Roth et al. 2021) anomaly map and detected defective region after thresholding are shown in Figure 5 (d). The Patchcore method can achieve successfully defective region localization. However, it cannot capture the detailed shape contour of the defective region. Therefore, its pixel-level defective region segmentation performance is not satisfactory.

The extracted background by RASL (Peng et al. 2012) is shown in Figure 5 (e). Since it can only deal with one channel image, we first convert the original images into greyscale. The image alignment of RASL requires an inset of the original images and we adopt the default setting where a 5-pixel inset is adopted. Therefore, the output decomposed anomaly image is $118 \times 118$. The tuning parameters are chosen by cross-validation. The extracted defective region is shown in Figure 5 (e). It fails to capture a large portion of the real defective region but instead brings in some noise from the background. This is because the low-rank (or low-rank after alignment) assumption does not apply to the background in this application. A relatively low dice coefficient of 59.1% is not surprising.

The reconstructed background by the proposed PAEDID method is shown in Figure 5 (f) which is close to the real background. This demonstrates that the patch-based approach can restrict the generalization capability of the AE while maintaining its modeling capability. The defective region extracted by the decomposition algorithm is shown in Figure 5 (f), which is close to the true defective region mask shown in Figure 5 (a). The dice coefficient of the proposed PAEDID method is 98.9%.

### *3.2 Simulation study for Product 2*

We also conduct a simulation study on product 2 of which the normal backgrounds are shown in Figure 1 (b). There are 400 normal images of dimension 600×600 and we resize them into 128×128. 350 images are randomly selected as training datasets in which 300 images are used as training and 50 images as the validation set to train the AE. The remaining 50 images are used to generate images with defects. Samples of the generated anomaly images with corresponding normal backgrounds and generated defective regions are shown in Figure 6 (a).

The encoder and encoder network share the same structure as the previous simulation study. The same tuning parameters $l, k, \alpha, \lambda$ and postprocessing step is adopted as in the previous simulation study. Since the images are aligned, we simplify the memory addressing algorithm by only searching for the closest patch in the memory bank of the same spatial location. We also compare the PAEDID method with SSIM AE (Bergmann et al. 2018), MemAE (Gong et al. 2019), PatchCore (Roth et al. 2021) and RASL (Peng et al. 2012). The qualitative result and quantitative result are shown in Figure 6 and Table 1, respectively. The proposed PAEDID outperforms all algorithms. Notice that the performance of RASL increases since the background variation across multiple images in product 2 is less than product 2 in the previous simulation study.

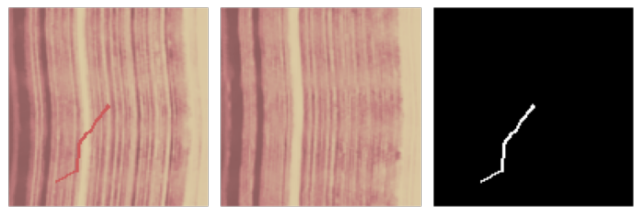

(a) Image with defect, true background, and simulated defective region

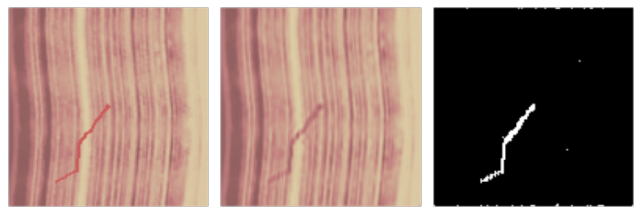

(b) Image with defect, SSIM AE reconstructed background and detected defective region

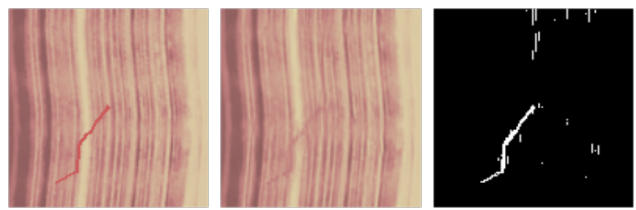

(c) Image with defect, MemAE reconstructed background and detected defective region

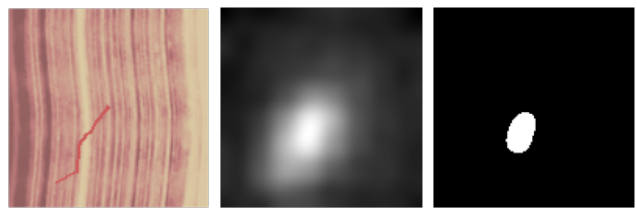

(d) Image with defect, PatchCore anomaly score map and detected defective region

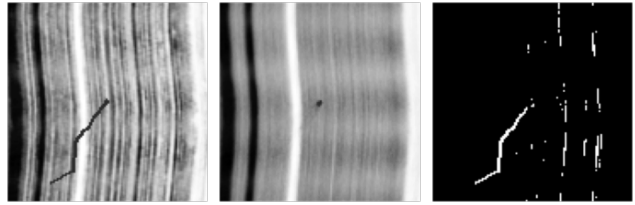

(e) Image with defect in greyscale, RASL reconstructed background and detected defective region

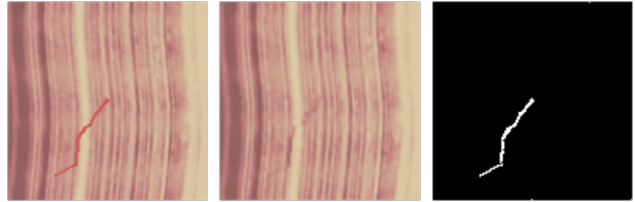

(f) Image with defect, PAEDID reconstructed background and detected defective region

Fig. 6. Comparison between the performance of SSIM AE, RASL PatchCore, and PAEDID

## 4. Case study

In this section, we apply the proposed PAEDID method on the ELPV dataset (Deitsch et al. 2021) containing (2624) grayscale images of normal (1508) and defective (1116) solar cells with varying degrees of degradation from different modules. Samples of normal and defective images are shown in Figure 7.

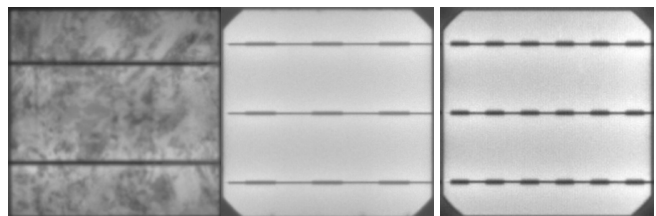

(a) normal images

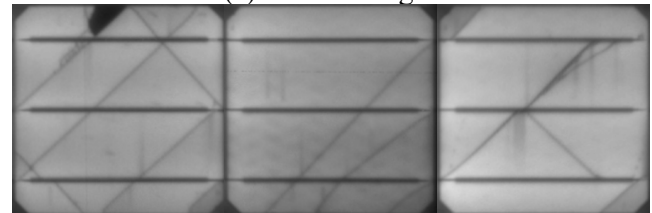

(b) images with defects

Fig. 7. Samples of different types of normal and defective solar cell images

It is easy to see that there are different types of normal images. It is hard to describe the normal image by either using the domain regularization type of statistical priors or an SSIM AE without a memory bank. The aim is to detect the crack type of defect. To demonstrate the defective region segmentation capability, we apply the PAEDID method on manually pixel-wise annotated 50 test images.

We first resize all images from dimension $300 \times 300$ into $128 \times 128$. In the proposed PAEDID method, there are 1508 images in the training dataset in which 1000 images are randomly chosen as training and 508 images are used as the validation set to train the AE. The encoder has 4 convolutional layers each of which is followed by a MaxPooling layer of pool size $2 \times 2$: Conv2D (128, 128, 3) - MaxPooling - Conv2D(64, 64, 9) - MaxPooling - Conv2D(32, 32, 27) - MaxPooling - Conv2D (16,16, 81) - MaxPooling. The decoder also has 4 convolutional layers each of which is followed by an UpSampling layer. The latent space is of shape $8 \times 8 \times 81$. Since the images are aligned, we simplify the memory addressing algorithm by only searching for the closest patch in the memory bank of the same spatial location.

For comparison, we also applied the other four state-of-the-art methods including SSIM AE (Bergmann et al. 2018), MemAE (Gong et al. 2019), PatchCore (Roth et al. 2021) and RASL (Peng et al. 2012). We assume that all input images are pre-identified defective images and evaluate the pixel-wise anomaly detection performance of the proposed method. Extracted defective region and reconstructed (or decomposed) background of one sample image by different methods are shown in Figure 8. The average dice coefficient (and standard deviation) of the proposed PAEDID method on the test set is reported in Table 2.

Table 2. Average dice coefficient (and standard deviation) for different methods in the case study

|  | Manually annotated defect (Dice coefficient) |
|---|---|
| SSIM AE | 14.2% (13.0%) |
| MemAE | 8.3% (4.4%) |
| PatchCore | 25.2% (8.7%) |
| RASL | 26.7% (9.2%) |
| PAE | 17.1% (12.1%) |
| PAEDID | 62.2% (14.9%) |

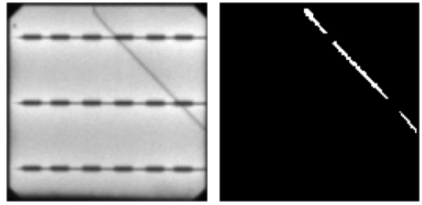

(a) Image with defect, label for defective region

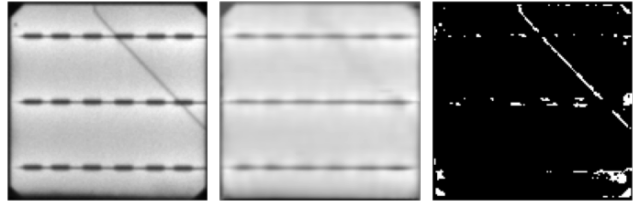

(b) Image with defect, SSIM AE reconstructed background and detected defective region

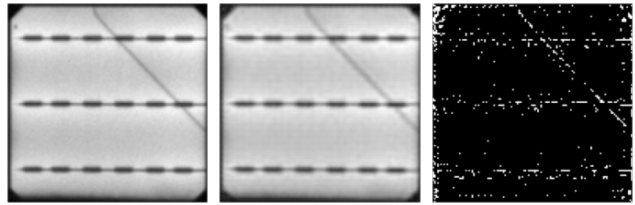

(c) Image with defect, MemAE reconstructed background and detected defective region

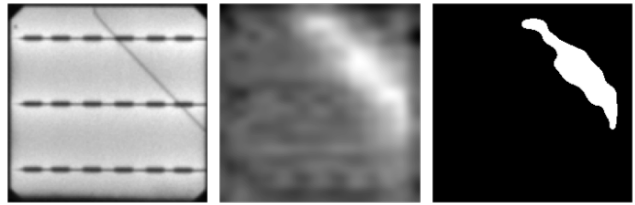

(d) Image with defect, PatchCore anomaly score map, and detected defective region

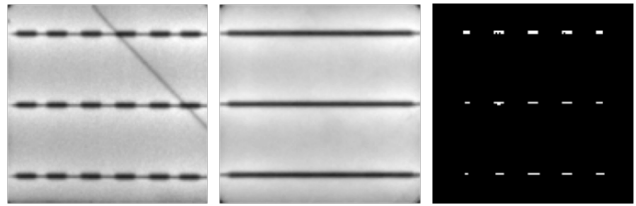

(e) Image with defect, RASL reconstructed background and detected defective region

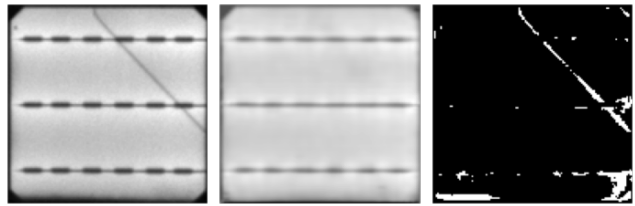

(f) Image with defect, PAE reconstructed background and detected defective region

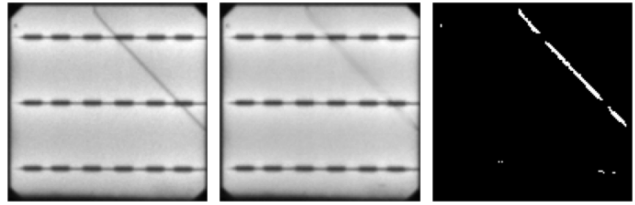

(g) Image with defect, PAEDID decomposed background, and detected defective region

Fig. 8. Comparison between the performance of SSIM AE, RASL, PatchCore, PAE, and PAEDID

Among those methods, the PAEDID method outperforms the other methods (SSIM AE (Bergmann et al. 2018), MemAE (Gong et al. 2019), PatchCore (Roth et al. 2021) and RASL (Peng et al. 2012)). The shape contour is almost perfectly extracted, except for the region where the crack intersects the dark line in the background. This is an inherent problem of all decomposition-based image processing algorithms without the additional assumption of the defective region, i.e., continuity, etc. For example, even if the background is reconstructed perfectly, the defective region cannot be extracted by subtracting the reconstructed background from the raw image. In this case, more advanced measurement technologies need to be adopted that can distinguish the defect and background.

In the PAEDID method, the tuning parameter $\lambda = 0.0006$ is chosen by cross-validation. The reconstructed background by PAE is used as an image prior which guides the PAEDID algorithm to decompose the raw image into the corresponding background and defects. It can capture a clear background and a high-quality defective region image as shown in Figure 8 (g). The improvement over the PAE method indicates that the deep image decomposition step is necessary which can further improve defective region segmentation performance.

*Ablation study of deep image decomposition step*

To demonstrate the necessity of the deep image decomposition step, we also report the performance of using PAE without the decomposition step, where a residual between the test image and its deep image prior is used for defective region segmentation. However, it can be seen from Figure 8 (b) (f) that both the reconstructed backgrounds by SSIM AE and PAE are blurry. Subsequently, sharp background features will also appear in the detected defects if directly subtracting the blurry reconstructed backgrounds from the raw image, which leads to the unsatisfactory performance of PAE and AE, with the average dice coefficients as PAE 17.1% and AE 14.2% respectively.

## 5. Conclusion

In this paper, we propose a novel unsupervised method for pixel-level defective region segmentation which combines the advantages of both deep-learning-based methods and matrix-decomposition-based methods. It is suitable for pixel-level defective region segmentation in complex images when there are limited defective samples and pixel-level labeling. The simulation and case studies demonstrate its superiority in challenging pixel-level defective region segmentation problems. One possible future direction is to adopt other types of generative models as deep prior for the normal images, which we leave for future work.

## Data Availability Statement

The data that support the findings of this study are openly available in (i) BTAD dataset (Mishra et al. 2021) at https://github.com/pankajmishra000/VT-ADL; (ii) ELPV dataset (Deitsch et al. 2021) at https://github.com/zae-bayern/elpv-dataset;

## References

Akcay, S., D. Ameln, A. Vaidya, B. Lakshmanan, N. Ahuja and U. Genc (2022). "Anomalib: A deep learning library for anomaly detection." *arXiv preprint arXiv:2202.08341*.

An, J. and S. Cho (2015). "Variational autoencoder based anomaly detection using reconstruction probability." *Special Lecture on IE* **2**(1): 1-18.

Baur, C., B. Wiestler, S. Albarqouni and N. Navab (2018). Deep autoencoding models for unsupervised anomaly segmentation in brain MR images. *International MICCAI brainlesion workshop*, Springer.

Bergmann, P., M. Fauser, D. Sattlegger and C. Steger (2019). MVTec AD--A comprehensive real-world dataset for unsupervised anomaly detection. *Proceedings of the IEEE/CVF Conference on Computer Vision and Pattern Recognition*.

Bergmann, P., M. Fauser, D. Sattlegger and C. Steger (2020). Uninformed students: Student-teacher anomaly detection with discriminative latent embeddings. *Proceedings of the IEEE/CVF Conference on Computer Vision and Pattern Recognition*.

Bergmann, P., S. Löwe, M. Fauser, D. Sattlegger and C. Steger (2018). "Improving unsupervised defect segmentation by applying structural similarity to autoencoders." *arXiv preprint arXiv:1807.02011*.

Bevandić, P., I. Krešo, M. Oršić and S. Šegvić (2022). "Dense open-set recognition based on training with noisy negative images." *Image and Vision Computing*: 104490.

Bouwmans, T. and E. H. Zahzah (2014). "Robust PCA via principal component pursuit: A review for a comparative evaluation in video surveillance." *Computer Vision and Image Understanding* **122**: 22-34.

Candès, E. J., X. Li, Y. Ma and J. Wright (2011). "Robust principal component analysis?" *Journal of the ACM (JACM)* **58**(3): 1-37.

Cen, J., P. Yun, J. Cai, M. Y. Wang and M. Liu (2021). Deep metric learning for open world semantic segmentation. *Proceedings of the IEEE/CVF International Conference on Computer Vision*.

Creusot, C. and A. Munawar (2015). Real-time small obstacle detection on highways using compressive RBM road reconstruction. *2015 IEEE Intelligent Vehicles Symposium (IV)*, IEEE.

Defard, T., A. Setkov, A. Loesch and R. Audigier (2021). Padim: a patch distribution modeling framework for anomaly detection and localization. *International Conference on Pattern Recognition*, Springer.

Deitsch, S., C. Buerhop-Lutz, E. Sovetkin, A. Steland, A. Maier, F. Gallwitz and C. Riess (2021). "Segmentation of photovoltaic module cells in uncalibrated electroluminescence images." *Machine Vision and Applications* **32**(4): 1-23.

Deng, J., W. Dong, R. Socher, L.-J. Li, K. Li and L. Fei-Fei (2009). Imagenet: A large-scale hierarchical image database. *2009 IEEE Conference on Computer Vision and Pattern Recognition*, IEEE.

Friedman, J. H. (2017). *The elements of statistical learning: Data mining, inference, and prediction*, springer open.

Gong, D., L. Liu, V. Le, B. Saha, M. R. Mansour, S. Venkatesh and A. v. d. Hengel (2019). Memorizing normality to detect anomaly: Memory-augmented deep autoencoder for unsupervised anomaly detection. *Proceedings of the IEEE/CVF International Conference on Computer Vision*.

Gudovskiy, D., S. Ishizaka and K. Kozuka (2022). Cflow-ad: Real-time unsupervised anomaly detection with localization via conditional normalizing flows. *Proceedings of the IEEE/CVF Winter Conference on Applications of Computer Vision*.

Hinton, G. E. and R. R. Salakhutdinov (2006). "Reducing the dimensionality of data with neural networks." *Science* **313**(5786): 504-507.

Kingma, D. P. and J. Ba (2014). "Adam: A method for stochastic optimization." *arXiv preprint arXiv:1412.6980*.

Lis, K., K. Nakka, P. Fua and M. Salzmann (2019). Detecting the unexpected via image resynthesis. *Proceedings of the IEEE/CVF International Conference on Computer Vision*.

Liu, L., W. Ouyang, X. Wang, P. Fieguth, J. Chen, X. Liu and M. Pietikäinen (2020). "Deep learning for generic object detection: A survey." *International journal of computer vision* **128**(2): 261-318.

Mardani, M., G. Mateos and G. B. Giannakis (2013). "Recovery of low-rank plus compressed sparse matrices with application to unveiling traffic anomalies." *IEEE Transactions on Information Theory* **59**(8): 5186-5205.

Minaee, S., A. Abdolrashidi and Y. Wang (2015). Screen content image segmentation using sparse-smooth decomposition. *2015 49th Asilomar Conference on Signals, Systems and Computers*, IEEE.

Minaee, S., Y. Y. Boykov, F. Porikli, A. J. Plaza, N. Kehtarnavaz and D. Terzopoulos (2021). "Image segmentation using deep learning: A survey." *IEEE Transactions on Pattern Analysis and Machine Intelligence*.

Mishra, P., R. Verk, D. Fornasier, C. Piciarelli and G. L. Foresti (2021). "VT-ADL: A vision transformer network for image anomaly detection and localization." *arXiv preprint arXiv:2104.10036*.

Mou, S. and J. Shi (2022). "Compressed smooth sparse decomposition." *arXiv preprint arXiv:2201.07404*.

Mou, S., A. Wang, C. Zhang and J. Shi (2021). "Additive tensor decomposition considering structural data information." *IEEE Transactions on Automation Science and Engineering*.

Pang, G., C. Shen, L. Cao and A. V. D. Hengel (2021). "Deep learning for anomaly detection: A review." *ACM Computing Surveys (CSUR)* **54**(2): 1-38.

Peng, Y., A. Ganesh, J. Wright, W. Xu and Y. Ma (2012). "RASL: Robust alignment by sparse and low-rank decomposition for linearly correlated images." *IEEE Transactions on Pattern Analysis and Machine Intelligence* **34**(11): 2233-2246.

Roth, K., L. Pemula, J. Zepeda, B. Schölkopf, T. Brox and P. Gehler (2021). "Towards total recall in industrial anomaly detection." *arXiv preprint arXiv:2106.08265*.

Schlegl, T., P. Seeböck, S. M. Waldstein, G. Langs and U. Schmidt-Erfurth (2019). "f-AnoGAN: Fast unsupervised anomaly detection with generative adversarial networks." *Medical image analysis* **54**: 30-44.

Schlegl, T., P. Seeböck, S. M. Waldstein, U. Schmidt-Erfurth and G. Langs (2017). Unsupervised anomaly detection with generative adversarial networks to guide marker discovery. *International Conference on InformationPprocessing in Medical Imaging*, Springer.

Simonyan, K. and A. Zisserman (2014). "Very deep convolutional networks for large-scale image recognition." *arXiv preprint arXiv:1409.1556*.

Wang, G., S. Han, E. Ding and D. Huang (2021a). "Student-teacher feature pyramid matching for unsupervised anomaly detection." *arXiv preprint arXiv:2103.04257*.

Wang, S., L. Wu, L. Cui and Y. Shen (2021b). Glancing at the patch: Anomaly localization with global and local feature comparison. *Proceedings of the IEEE/CVF Conference on Computer Vision and Pattern Recognition*.

Wang, Z., A. C. Bovik, H. R. Sheikh and E. P. Simoncelli (2004). "Image quality assessment: from error visibility to structural similarity." *IEEE Transactions on Image Processing* **13**(4): 600-612.

Xu, H., C. Caramanis and S. Sanghavi (2010). "Robust PCA via outlier pursuit." *arXiv preprint arXiv:1010.4237*.

Xu, H., C. Caramanis and S. Sanghavi (2012). "Robust PCA via outlier pursuit." *IEEE Transactions on Information Theory* **58**(5): 3047-3064.

Yan, H., K. Paynabar and J. Shi (2017). "Anomaly detection in images with smooth background via smooth-sparse decomposition." *Technometrics* **59**(1): 102-114.

Yan, H., K. Paynabar and J. Shi (2018). "Real-time monitoring of high-dimensional functional data streams via spatio-temporal smooth sparse decomposition." *Technometrics* **60**(2): 181-197.

Yu, J., Y. Zheng, X. Wang, W. Li, Y. Wu, R. Zhao and L. Wu (2021). "FastFlow: Unsupervised anomaly detection and localization via 2D normalizing flows." *arXiv preprint arXiv:2111.07677*.

Zenati, H., C. S. Foo, B. Lecouat, G. Manek and V. R. Chandrasekhar (2018). "Efficient gan-based anomaly detection." *arXiv preprint arXiv:1802.06222*.

Zhao, S., J. Song and S. Ermon (2017). "Towards deeper understanding of variational autoencoding models." *arXiv preprint arXiv:1702.08658*.

Zhou, C. and R. C. Paffenroth (2017). Anomaly detection with robust deep autoencoders. *Proceedings of the 23rd ACM SIGKDD International Conference on Knowledge Discovery and Data Mining*.

Zou, K. H., S. K. Warfield, A. Bharatha, C. M. Tempany, M. R. Kaus, S. J. Haker, W. M. Wells III, F. A. Jolesz and R. Kikinis (2004). "Statistical validation of image segmentation quality based on a spatial overlap index1: scientific reports." *Academic Radiology* **11**(2): 178-189.

**Appendix A**

In this section, we introduce the SSIM loss function. SSIM loss refers to structural similarity loss (Wang et al. 2004), which is a perceptual metric to measure the similarity of two images. It is defined as

$$\|x - y\|_{SSIM} = \frac{(2\mu_x\mu_y + c_1)(2\sigma_{xy} + c_2)}{(\mu_x^2 + \mu_y^2 + c_1)(\sigma_x^2 + \sigma_y^2 + c_2)}$$

where $\mu_x$ and $\mu_y$ indicate the pixel sample mean of $x$ and $y$; $\sigma_x^2$ and $\sigma_y^2$ indicates the variance of $x$ and $y$; $\sigma_{xy}$ indicates the correlation between $x$ and $y$; $c_1$ and $c_2$ are constants whose value can be found in Wang et al. (2004). Using SSIM loss and has demonstrated superior performance in AE-based anomaly detection tasks (Bergmann et al. 2018).

**Appendix B**

In this section, we introduce the decomposition framework considering measurement noise. We assume that any test image is a superimposition of (i) normal image (background $L$) from the distribution learned by the AE, and (ii) sparse defective regions ($S$) and (iii) element-wise independent Gaussian measurement noise $e$, i.e., $X = L + S + e$. Then, the defective region segmentation can be formulated as a matrix decomposition problem with a combination of a deep image prior for the background and sparsity regularization for the defective regions. Given a test image $X$, we aim to extract the defective regions by solving the following penalized optimization problem:

$$\min_{L,S} \|X - L - S\|_2^2 + \lambda_1 \|S\|_1 + \lambda_2 \|L - \hat{X}\|_{SSIM} \qquad (4)$$

where $\lambda_1$ and $\lambda_2$ are tuning parameters.

**Appendix C**

In some applications, the memory bank size can be large. The following algorithm can be used for reducing the memory bank size.

| |
|---|
| **Algorithm 2.** Algorithm for greedy coreset selection |
| Input $\mathcal{M}_{agg,l}^{mat}$, coreset size $n_c$. |
| Initialize $\mathcal{M}_{agg,l}^{mat,C} = \{\}$. |

**For** $i \in [0, \ldots, n_c - 1]$ **do:**
$$m_i = \arg \max_{m \in \mathcal{M}_{agg,l} \setminus \mathcal{M}_{agg,l}^C} \min_{n \in \mathcal{M}_{agg,l}^C} \|\psi(m) - \psi(n)\|_2$$
$$\mathcal{M}_{agg,l}^{mat,C} \leftarrow \mathcal{M}_{agg,l}^{mat,C} \cup m_i$$
**End**
$$\mathcal{M}_{agg,l}^{mat} \leftarrow \mathcal{M}_{agg,l}^{mat,C}$$

where $\psi(\cdot)$ is a random linear projection that projects the input feature to a lower dimensional space. We set it as 128 for all simulation studies and case studies.